\documentclass[12pt,a4paper]{article}
\usepackage[a4paper,left=2.5cm,right=2.5cm,top=2.5cm,bottom=2.5cm]{geometry}
\usepackage{newtxtext,newtxmath}
\usepackage{amsmath,bm}
\usepackage{graphicx}
\usepackage{booktabs,longtable,tabularx,array,calc}
\usepackage{multirow}
\usepackage{ragged2e}
\usepackage{setspace}
\usepackage{indentfirst}
\usepackage{hyperref}
\usepackage{caption}
\usepackage{float}
\usepackage{etoolbox}
\usepackage{microtype}
\usepackage{enumitem}
\graphicspath{{media/}}
\hypersetup{hidelinks,pdftitle={Topology-Prior-Guided Regional Latent Dynamics for Transient Structural Response Prediction}}

\usepackage{needspace}
\usepackage[section]{placeins}
\renewcommand{\textsubscript}[1]{\raisebox{-.3ex}{\scriptsize #1}}
\begin{document}
\sloppy
\emergencystretch=3em

\begin{center}{\large\bfseries Topology-Prior-Guided Regional Latent Dynamics for Transient Structural Response Prediction\par}\vspace{0.85em}

Yuan Zhang\textsuperscript{1}, Mohammad Amin Nabian\textsuperscript{2}, ShaoTong Fu\textsuperscript{2}, YiDing Liu\textsuperscript{1}, ChengCheng Chu\textsuperscript{1},XueLin Tao\textsuperscript{1},Hua Li\textsuperscript{1,*}\par\vspace{0.35em}

\textsuperscript{1}TSYG AI Corporation\par\vspace{0.35em}

\textsuperscript{2}NVIDIA Corporation\par\vspace{0.35em}

{\small zhangyuan@tsygai.com; mnabian@nvidia.com; shaotongf@nvidia.com; liuyiding@tsygai.com; chuchengcheng@tsygai.com; taoxuelin@tsygai.com; lihua@tsygai.com}\par\vspace{0.35em}

\end{center}\vspace{0.6em}

\noindent\textbf{Abstract}\par

Predicting the transient nonlinear responses of complex engineering structures remains challenging due to large finite element meshes, heterogeneous structural components, and history-dependent mechanical behavior. This study proposes a topology-prior-guided regional latent dynamics (TPLD) surrogate model that combines physically structured regional representations with dynamics-aware latent learning for full-field time-history prediction. Based on mesh topology and engineering priors, structural responses are encoded into spatially coupled regional latent variables through region-aware feature extraction and topology-constrained interactions. An auxiliary one-step prediction task enhances latent-space dynamical predictability by introducing state-transition supervision. A load-conditioned spatiotemporal Mamba model then predicts regional latent trajectories by integrating temporal causal scanning and bidirectional spatial scanning, while a shared coordinate-conditioned decoder reconstructs full-field nodal responses. The framework is evaluated on three finite element datasets: a parametric stiffened panel, a partitioned pressure impact plate, and a Toyota Yaris full-vehicle crash model. The relative L2 errors for the three displacement components range from 4.00\% to 8.21\% on the stiffened panel and from 1.41\% to 12.22\% on the approximately 1.49-million-node vehicle model. On the BIW crash benchmark, TPLD achieves a relative L2 position error of 0.620\%, representing a 30.7\% reduction compared with the best-performing GeoTransolver configuration. Ablation studies demonstrate the advantages of topology-guided regional representations over global latent and learnable soft-slice alternatives, while auxiliary one-step prediction particularly improves the prediction of accumulated plastic strain.


\section{Introduction}\label{sec:1}

Complex engineering structures exhibit nonlinear responses to impact, collision, and other transient loads, including global deformation, localized stress concentrations, and plastic strain accumulation. Although high-fidelity finite element (FE) analysis can resolve these responses while accounting for complex geometries and material nonlinearities, its computational cost increases substantially with structural complexity, mesh resolution, and simulation duration. Repeated analyses for structural optimization, parametric studies, and varying loading conditions therefore motivate the development of efficient surrogate models capable of predicting full-field response histories.

Reduced-order modeling offers a means of alleviating this computational cost. Conventional approaches, such as proper orthogonal decomposition (POD), approximate solutions using fixed linear bases, which can be restrictive for strongly nonlinear responses (\hyperlink{ref:benner2015}{Benner et al., 2015}). Deep autoencoders have been used to learn nonlinear low-dimensional manifolds (\hyperlink{ref:lee2020}{Lee and Carlberg, 2020}), and deep learning-based reduced-order models have been developed for time-dependent parametric problems (\hyperlink{ref:fresca2021}{Fresca et al., 2021}; \hyperlink{ref:fresca2022}{Fresca and Manzoni, 2022}). For transient structural dynamics, autoencoders have been combined with recurrent or feedforward predictors (\hyperlink{ref:simpson2021}{Simpson et al., 2021}; \hyperlink{ref:nikolopoulos2022}{Nikolopoulos et al., 2022}), while latent-space dynamics identification offers another strategy for efficient simulation (\hyperlink{ref:fries2022}{Fries et al., 2022}). Related studies have explored representations better suited to dynamical evolution through Koopman embeddings and jointly learned latent coordinates and governing equations (\hyperlink{ref:lusch2018}{Lusch et al., 2018}; \hyperlink{ref:champion2019}{Champion et al., 2019}). Nevertheless, accurate reconstruction of response fields at individual time steps does not necessarily ensure that the learned latent states are suitable for predicting subsequent responses. Moreover, global latent representations may obscure localized behavior in structures with heterogeneous components. These considerations motivate the development of latent representations that combine spatial expressiveness with dynamical predictability.

Graph-based learning provides flexible tools for extracting spatial features from unstructured FE meshes. Graph networks and MeshGraphNets have demonstrated the value of message passing for learning physical interactions on irregular discretizations (\hyperlink{ref:sanchez2020}{Sanchez-Gonzalez et al., 2020}; \hyperlink{ref:pfaff2021}{Pfaff et al., 2021}; \hyperlink{ref:brandstetter2022}{Brandstetter et al., 2022}). Their applications to structural dynamics include full-field response prediction and multihierarchical surrogate modeling (\hyperlink{ref:li2023}{Li Q et al., 2023}; \hyperlink{ref:kneifl2025}{Kneifl et al., 2025}). Geometry-informed and FE-informed neural operators have likewise been investigated for complex geometries and structural responses (\hyperlink{ref:liz2023}{Li Z et al., 2023}; \hyperlink{ref:yang2025}{Yang et al., 2025}), alongside spatiotemporal graph operators and heterogeneous GNNs for nonlinear FE problems (\hyperlink{ref:du2026}{Du et al., 2026}; \hyperlink{ref:li2025}{Li et al., 2025}). These studies demonstrate the importance of mesh connectivity, geometric information, and spatial interactions in learning structural responses. However, for reduced latent dynamics, organizing these features into compact representations with explicit correspondence to physical structural regions remains a distinct challenge.

Attention-based aggregation offers a complementary approach to compressing large sets of nodal features. Set Transformer and Perceiver IO employ attention and latent queries to aggregate information from unstructured inputs (\hyperlink{ref:lee2019}{Lee J et al., 2019}; \hyperlink{ref:jaegle2022}{Jaegle et al., 2022}), while Transolver constructs learnable physical slices to capture interactions over complex geometries (\hyperlink{ref:wu2024}{Wu et al., 2024}). Although these approaches provide flexible latent representations, learned aggregations do not necessarily establish explicit correspondence with physically defined structural components. For engineering structures comprising heterogeneous materials, varying section properties, and complex connections, incorporating engineering priors into regional latent-space organization may help preserve local response characteristics while supporting interactions among connected regions.

In addition to spatial representation, transient response prediction requires modeling loading-history dependence and dynamical interactions among structural regions. State-space models (SSMs), including S4 and Mamba, provide mechanisms for learning temporal dependencies through structured or selective state updates (\hyperlink{ref:gu2021}{Gu et al., 2021}; \hyperlink{ref:gu2023}{Gu and Dao, 2023}; \hyperlink{ref:dao2024}{Dao and Gu, 2024}). Bidirectional state-space modeling and two-dimensional selective scanning have also been explored for spatial representation learning (\hyperlink{ref:zhu2024}{Zhu et al., 2024}; \hyperlink{ref:liu2024}{Liu et al., 2024}). These developments suggest the potential of combining temporal state-space modeling with spatial information exchange. However, structural dynamics involves the coupled effects of external loading histories, region-specific responses, and structural connectivity. Capturing these effects requires a latent dynamics model that accounts for both temporal evolution and interactions among physically organized structural regions.

To address these challenges, this study proposes a topology-prior-guided regional latent dynamics (TPLD) surrogate model for full-field time-history prediction of transient structural responses. The framework integrates engineering-prior-guided regional representation learning, auxiliary dynamical supervision, and load-conditioned spatiotemporal latent evolution. The main contributions are summarized as follows:

(1) A topology-prior-guided regional latent representation learning method is developed. FE nodes are organized into physically meaningful regions using mesh topology and engineering information, including structural components, material properties, section characteristics, and loading and boundary conditions. Intra-region graph message passing, attention-based pooling, and topology-constrained inter-region interactions produce compact regional latent representations with explicit structural correspondence, preserving local response features and inter-region coupling information.

(2) An auxiliary one-step prediction mechanism is introduced to improve latent-space dynamical predictability. By predicting subsequent regional latent states from consecutive states and corresponding loading information, the auxiliary task provides additional supervision of dynamical transitions. Joint optimization in the latent space and physical response space encourages the learned representations to retain information relevant to temporal evolution rather than response-field reconstruction alone.

(3) A load-conditioned spatiotemporal Mamba latent dynamics model is developed for full-field time-history prediction. Temporal causal scanning captures loading-history dependence, while bidirectional spatial scanning models interactions among structural regions. The predicted regional latent trajectories are subsequently decoded into high-dimensional nodal response fields through a shared coordinate-conditioned decoder.

The proposed framework is evaluated on three structural dynamics datasets involving varying panel geometries, spatially nonuniform pressure loading, and full-vehicle frontal impacts. Its predictive performance is further compared with GeoTransolver variants on a BIW automotive crash benchmark, while ablation experiments investigate the contributions of topology-guided regional representations and auxiliary one-step prediction.

\section{Methodology}\label{sec:2}

\subsection{Framework Overview}\label{sec:2.1}

\begin{figure}[!htb]\centering
\includegraphics[width=0.96\linewidth,height=0.57\textheight,keepaspectratio]{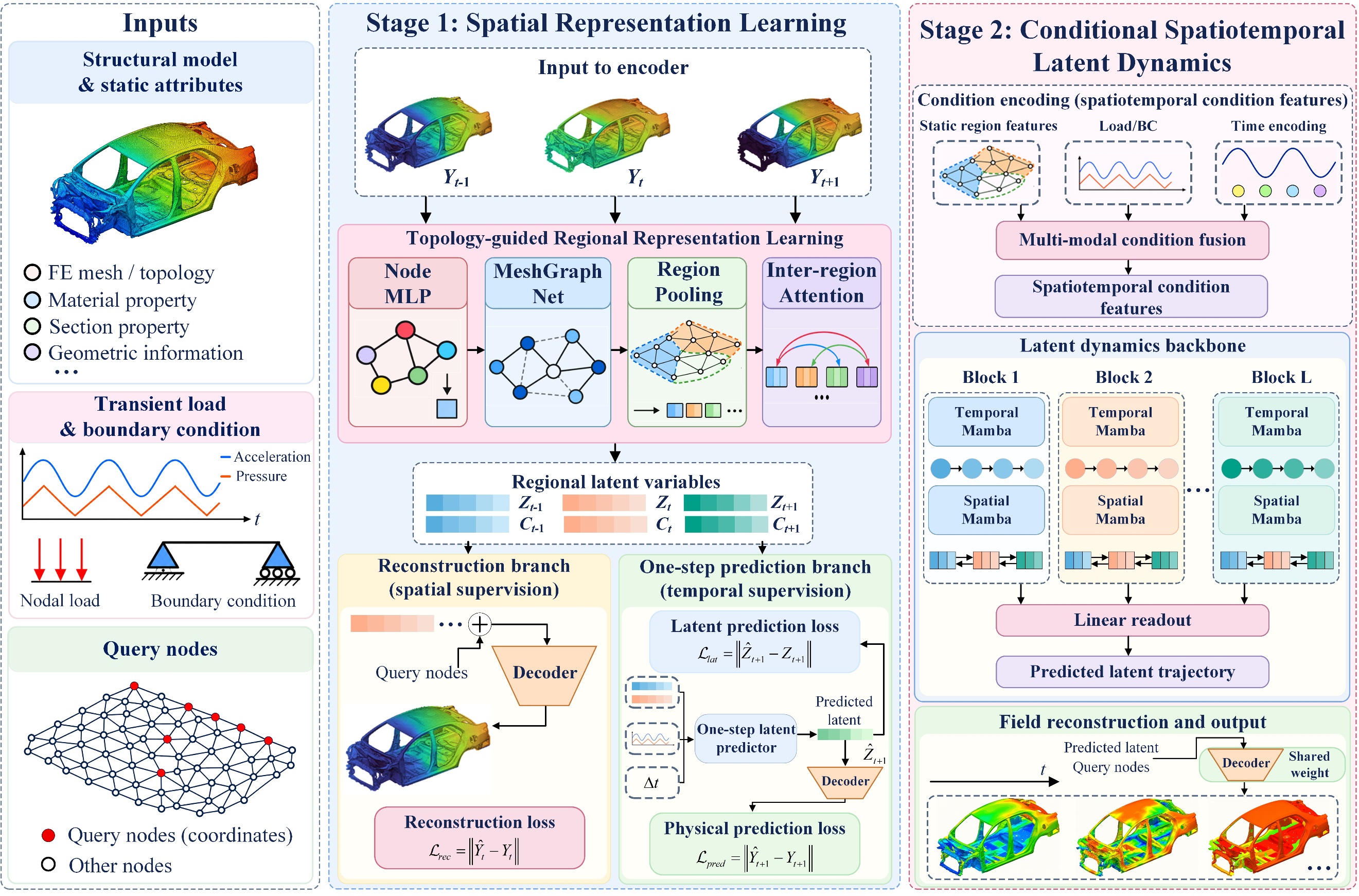}
\caption{Overall architecture of the proposed TPLD framework.}\label{fig:1}
\end{figure}

To enable efficient full-field time-history prediction of complex structural responses, this study proposes a topology-prior-guided regional latent dynamics (TPLD) surrogate model, as illustrated in Fig. \hyperref[fig:1]{1}. The framework combines physically organized regional latent representations with load-conditioned dynamical modeling to preserve local response characteristics and inter-region coupling while improving latent-space dynamical predictability.

Consider a structural system discretized by a finite element (FE) mesh, represented as a graph G = (V, E), where V and E denote the sets of nodes and mesh edges, respectively, and N is the total number of nodes. The structural response field at time t is denoted by \ensuremath{\mathbf{Y}_t\in\mathbb{R}^{N\times d_y}} , where  \ensuremath{d_y}  is the number of response variables. The static structural descriptors and prescribed loading history are denoted by \ensuremath{\mathbf{X}} and \ensuremath{\mathbf{L}_0}  respectively. Based on engineering priors and mesh topology, the nodes are partitioned into M regions, denoted by  \ensuremath{\mathcal{P}=\{V_1,V_2,\ldots,V_M\}} . The objective is to construct regional latent representations, predict their temporal evolution under prescribed loading conditions, and recover the corresponding full-field nodal responses.

The TPLD framework comprises two main components: regional latent representation learning and latent-space dynamics prediction. In the first component, intra-region graph message passing and attention-based pooling extract compact regional latent variables, while topology-constrained inter-region interactions provide contextual information among connected regions. A shared coordinate-conditioned decoder reconstructs nodal response fields from the regional representations. \Needspace{125pt}
The encoding and reconstruction processes are expressed as

\begin{equation}\label{eq:1}
\mathbf{Z}_t = \mathcal{E}_{\theta}(\mathbf{Y}_t;G,\mathbf{X},\mathcal{P})
\end{equation}

\begin{equation}\label{eq:2}
\widetilde{\mathbf{Y}}_t = \mathcal{D}_{\psi}(\mathbf{Z}_t;\mathbf{X},\mathcal{P})
\end{equation}

Where \ensuremath{\mathcal{E}_{\theta}} and \ensuremath{\mathcal{D}_{\psi}} denote the regional spatial encoder and coordinate-conditioned decoder, respectively.  \ensuremath{\mathbf{Z}_t\in\mathbb{R}^{M\times d_z}} represents the collection of regional latent variables,  \ensuremath{d_z} is the latent dimension of each region, and \ensuremath{\widetilde{\mathbf{Y}}_t} denotes the reconstructed response field. To enhance dynamical predictability, an auxiliary one-step prediction task is incorporated during representation learning, providing joint supervision of latent-state transitions and their corresponding physical response fields.

In the second component, a load-conditioned spatiotemporal Mamba model predicts regional latent trajectories using temporal causal scanning and bidirectional spatial scanning to capture loading-history dependence and inter-region dynamical coupling, respectively. The predicted latent trajectories are subsequently mapped to full-field nodal responses through the shared decoder:

\begin{equation}\label{eq:3}
\widehat{\mathbf{Z}}_{0:T} = \mathcal{F}_{\phi}(\mathbf{L}_{0:T},\Delta t_{1:T},\mathbf{X}_R,G_R)
\end{equation}

\begin{equation}\label{eq:4}
\widehat{\mathbf{Y}}_t = \mathcal{D}_{\psi}(\widehat{\mathbf{Z}}_t;\mathbf{X},\mathcal{P}),\quad t=0,\ldots,T
\end{equation}

where \ensuremath{\mathcal{F}_{\phi}} denotes the latent-space dynamics model,  \ensuremath{G_R} represents the regional adjacency topology,  \ensuremath{\mathbf{X}_R} contains the static regional descriptors, and \ensuremath{\Delta t_1} denotes the sequence of time increments.

Training proceeds in three stages. Stage 1 jointly optimizes the spatial encoder and decoder through response-field reconstruction and auxiliary one-step prediction. Stage 2 freezes both modules and trains the spatiotemporal dynamics model using reference latent trajectories encoded from FE response fields. Stage 3 keeps the spatial encoder frozen while jointly fine-tuning the dynamics model and shared decoder using physical-field prediction errors.

\FloatBarrier
\subsection{Topology-guided regional latent representation learning}\label{sec:2.2}

\begin{figure}[!htb]\centering
\includegraphics[width=0.96\linewidth,height=0.57\textheight,keepaspectratio]{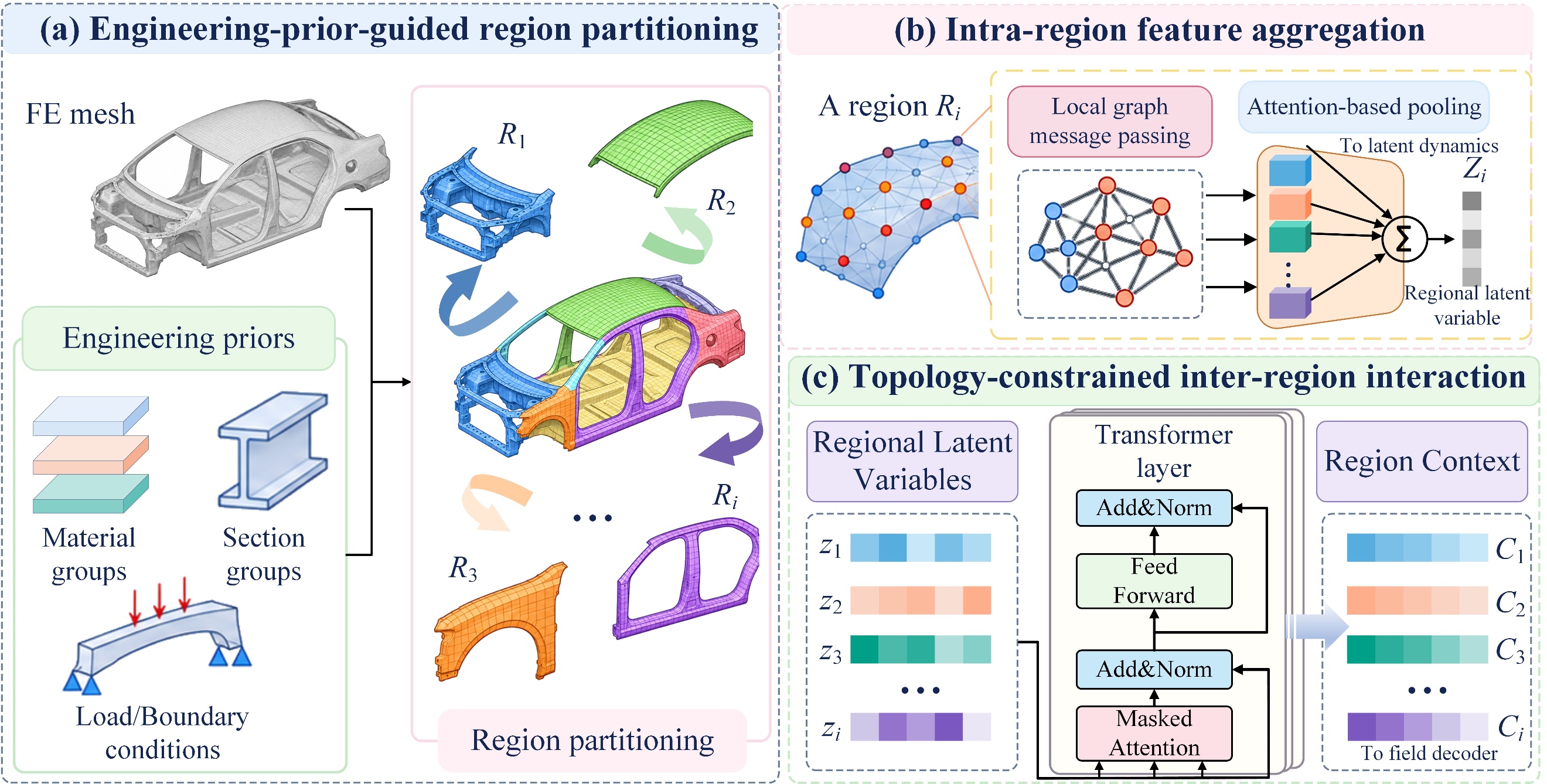}
\caption{Topology-guided regional latent representation learning}\label{fig:2}
\end{figure}

To construct a low-dimensional latent space that captures both local response characteristics and structural coupling, this study adopts an engineering-prior-guided regional representation learning approach, as illustrated in Fig. \hyperref[fig:2]{2}. First, the FE nodes are partitioned into regions based on engineering information, including structural components, geometric characteristics, and loading and boundary conditions. A regional adjacency graph is then constructed according to mesh connectivity. The partitioning criteria are tailored to each structure to establish correspondence between the resulting regions and physically meaningful structural features. Edges connecting nodes within the same region are retained to form intra-region graphs, on which MeshGraphNet (MGN) performs local message passing. The nodal responses and static features are first projected into a hidden feature space. Multiple layers of edge updates, neighborhood message aggregation, and node updates are then applied to extract spatial features within each region:

\begin{equation}\label{eq:5}
\mathbf{H}_t = \operatorname{MGN}([\mathbf{Y}_t;\mathbf{X}],G_{\mathrm{intra}},\mathbf{A}_e)
\end{equation}

where \ensuremath{\mathbf{H}_t} denotes the updated nodal features,  \ensuremath{G_{\mathrm{intra}}} represents the intra-region mesh graph, and \ensuremath{\mathbf{A}_e} denotes the set of edge features.

To accommodate regions with different numbers of nodes, an attention-based pooling mechanism inspired by set feature aggregation (\hyperlink{ref:lee2019}{Lee J et al., 2019}) is employed to compress the nodal features within each region into a fixed-dimensional latent variable. Specifically, query vectors are generated according to region types and used to compute attention weights with the key vectors of nodes within each region. The weights are normalized using Softmax within each region and subsequently applied to aggregate the node value features:

\begin{equation}\label{eq:6}
\mathbf{z}_{t,r} = f_p\!\left(\sum_{i\in V_r}\alpha_{t,i}\mathbf{v}_{t,i}\right)
\end{equation}

where  \ensuremath{\alpha_{t,i}}  denotes the attention weight normalized within the corresponding region,  \ensuremath{\mathbf{v}_{t,i}} is the node value vector,  \ensuremath{f_p} represents a nonlinear mapping function, and \ensuremath{\mathbf{z}_{t,r}} denotes the latent variable of region r. The attention weights characterize the contributions of individual nodes to the regional representation. The latent variables of all regions collectively form the regional latent representation at time t.

To account for interactions among structural regions, a topology-constrained regional attention mechanism is further introduced. Static regional information, including geometric characteristics and structural types, is first embedded into the latent feature space. Multi-head attention is then employed to exchange information among regions. The regional adjacency graph defines an attention mask that restricts each region to attending to itself and its neighboring regions. The resulting regional context representation is expressed as

\begin{equation}\label{eq:7}
\mathbf{C}_t = \mathbf{Z}_t + f_{\mathrm{int}}\!\left(\mathbf{Z}_t + f_{\mathrm{stat}}(\mathbf{S}),\mathbf{A}_R\right)
\end{equation}

where \ensuremath{\mathbf{S}} denotes the static regional information,  \ensuremath{f_{\mathrm{stat}}} is the static feature mapping function,  \ensuremath{\mathbf{A}_R} represents the regional adjacency mask, and \ensuremath{f_{\mathrm{int}}} denotes the regional attention interaction and output mapping.  \ensuremath{\mathbf{C}_t} is the resulting regional context representation. The regional latent variables are used for subsequent dynamical evolution modeling, while the regional context representations provide additional information for nodal response-field reconstruction.

A coordinate-conditioned implicit decoder is employed to reconstruct the nodal response fields. The local coordinates of query nodes are encoded using Fourier feature mapping (\hyperlink{ref:tancik2020}{Tancik et al., 2020}) and combined with nodal static features as inputs to a multilayer perceptron (MLP). In addition, feature-wise linear modulation (FiLM) (Perez E et al., 2018) is used to condition the decoding process. The latent variable, context representation, and static information associated with each node's region are used to generate modulation parameters. The reconstructed nodal response is expressed as

\begin{equation}\label{eq:8}
\mathbf{y}_{t,i} = f_{\psi}\!\left(\gamma(\boldsymbol{\xi}_i),\mathbf{x}_i;\mathbf{z}_{t,r(i)},\mathbf{c}_{t,r(i)},\mathbf{s}_{r(i)}\right)
\end{equation}

Where  \ensuremath{\boldsymbol{\xi}_i} denotes the local coordinates of node i,  \ensuremath{\gamma} represents the Fourier feature encoding,  \ensuremath{\mathbf{x}_i} denotes the nodal static features, and  \ensuremath{r(i)}  identifies the region to which node i belongs. The terms  \ensuremath{\mathbf{c}_{t,r(i)}}  and  \ensuremath{\mathbf{s}_{r(i)}}  represent the corresponding regional context and static information, respectively, while  \ensuremath{f_{\psi}}  denotes the FiLM-conditioned decoding network.

For nodes with prescribed values, boundary conditions are further enforced at the decoder output. Through these operations, the proposed method establishes a mapping from high-dimensional nodal response fields to regional latent variables and reconstructs the structural response fields using regional latent representations and nodal spatial information. The resulting regional latent space provides the spatial representation required for subsequent dynamical prediction.

\FloatBarrier
\subsection{Dynamics-aware latent learning and spatiotemporal prediction}\label{sec:2.3}

\begin{figure}[!htb]\centering
\includegraphics[width=0.96\linewidth,height=0.57\textheight,keepaspectratio]{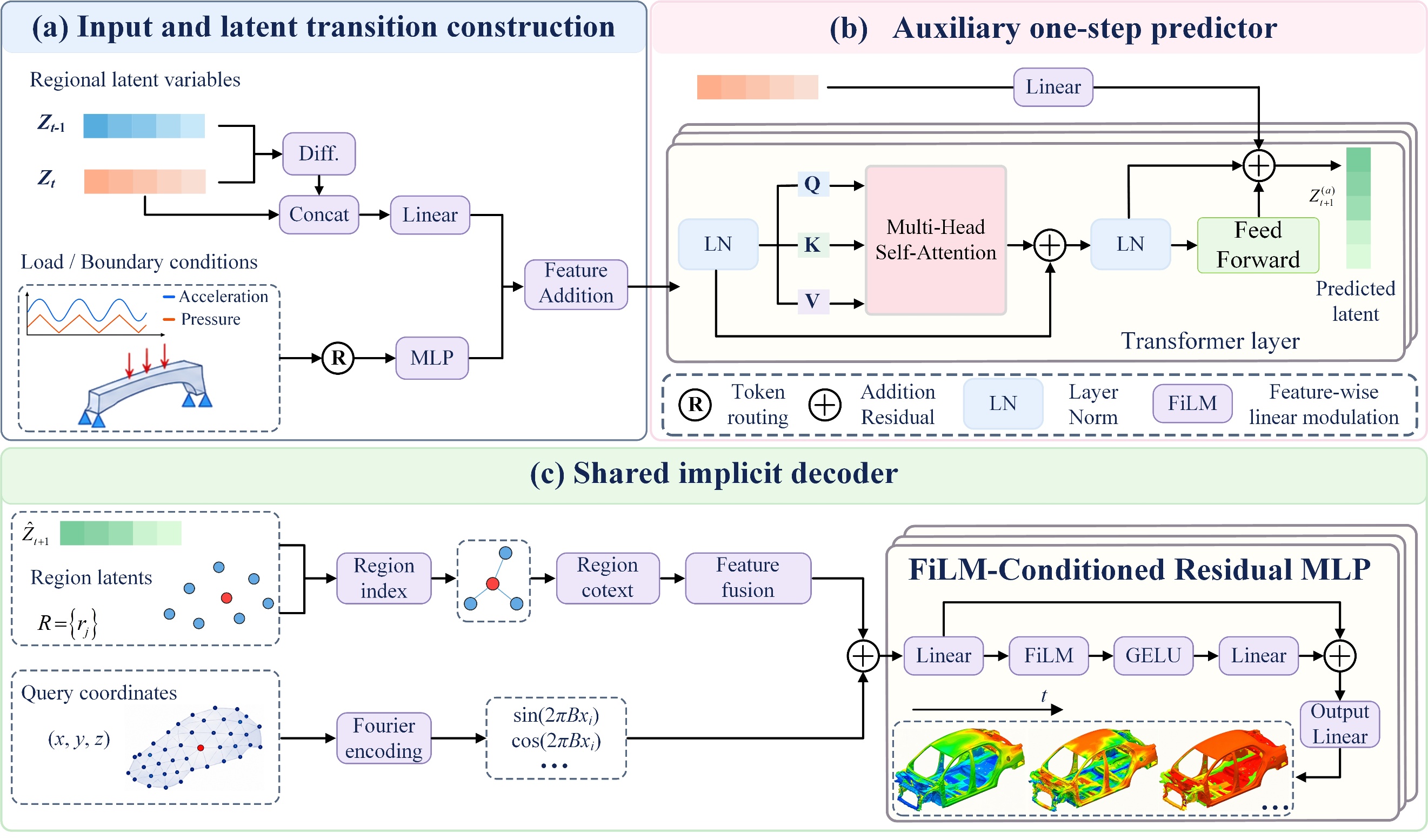}
\caption{Auxiliary one-step prediction and shared response field decoding}\label{fig:3}
\end{figure}

Training the spatial encoder solely through response-field reconstruction does not guarantee that the learned latent variables retain sufficient information for dynamical evolution. To address this limitation, an auxiliary one-step prediction task is introduced during spatial representation learning, as illustrated in Fig. \hyperref[fig:3]{3}. This task predicts the subsequent latent state using regional latent variables from two consecutive time steps, together with the corresponding loading and temporal information. Specifically, ground-truth response fields at three consecutive time steps are encoded using the shared encoder to obtain regional latent variables \ensuremath{\mathbf{Z}_{t-1}} , \ensuremath{\mathbf{Z}_t} ,and \ensuremath{\mathbf{Z}_{t+1}} . The auxiliary predictor takes the current latent variables and their temporal changes as inputs, together with external excitation and time increments. A regional Transformer is employed to capture state transitions and inter-region coupling, and the subsequent regional latent variables are predicted through a residual update:

\begin{equation}\label{eq:9}
\mathbf{Z}^{(a)}_{t+1} = \mathbf{Z}_t + \frac{\Delta t_t}{\Delta t_{\mathrm{ref}}}f_a\!\left(\mathbf{Z}_t,\Delta\mathbf{Z}_t,\mathbf{L}_t,\mathbf{L}_{t+1},\Delta t_t;G_R,\mathbf{X}_R\right)
\end{equation}

where  \ensuremath{\mathbf{Z}_{t+1}^{(a)}} denotes the regional latent variables predicted by the auxiliary predictor,  \ensuremath{f_a} represents the auxiliary prediction network, and \ensuremath{\Delta t_{\mathrm{ref}}} is the reference time-increment scale. The predictor updates the latent state \ensuremath{\mathbf{Z}_t} through a residual formulation.

The predicted latent variables are subsequently passed through the shared coordinate-conditioned decoder to reconstruct the response field at the next time step. During training, the latent variables obtained by encoding the ground-truth response field at the next time step serve as supervision targets, with gradient propagation stopped through the target branch. In addition, physical-field supervision is imposed on the response field decoded from the predicted latent variables. This joint supervision encourages the latent-state transitions to maintain consistency with the target latent variables while preserving the ability to reconstruct the corresponding nodal response distributions. The joint optimization objective for Stage 1 is formulated as

\begin{equation}\label{eq:10}
\mathcal{L}_{\mathrm{S1}} = \mathcal{L}_{\mathrm{rec}}+\mathcal{L}_{\mathrm{str}} + \lambda_{z}^{(a)}\mathcal{L}_{z}^{(a)}+\lambda_{y}^{(a)}\mathcal{L}_{y}^{(a)}
\end{equation}

where  \ensuremath{\mathcal{L}_{\mathrm{rec}}} denotes the response-field reconstruction loss at the current and subsequent time steps, and \ensuremath{\mathcal{L}_{\mathrm{str}}} is a weighted structural constraint loss incorporating mesh-edge differences and spatial gradient terms. The auxiliary latent prediction loss \ensuremath{\mathcal{L}^{(a)}_z} is defined as the mean squared error between the predicted latent variables and the ground-truth latent variables with stopped gradients. The auxiliary physical-field prediction loss  \ensuremath{\mathcal{L}^{(a)}_y}  measures the discrepancy between the decoded prediction and the ground-truth response field at the next time step. The corresponding \ensuremath{\lambda} terms denote the loss weights.

Building on the learned regional latent representation, a spatiotemporal latent dynamics model is constructed by integrating temporal causal scanning and bidirectional spatial scanning, as shown in Stage 2 of Fig. \hyperref[fig:1]{1}. First, static structural information, external loading histories, and temporal encodings are mapped into spatiotemporal conditioning features. Load information is then routed to the corresponding structural regions based on the predefined load–region associations, providing region-specific conditioning for the subsequent dynamics modeling. Subsequently, spatiotemporal Mamba blocks are employed to learn the dynamical evolution of regional latent variables. The temporal Mamba performs causal scanning along the time dimension, using selective state-space updates to capture dependencies on historical excitation and latent states. For spatial scanning, regions are ordered according to the Morton (Z-order) codes of their geometric centroids. Bidirectional scans are then applied to the resulting sequence to model inter-region interactions. These two scanning mechanisms jointly capture the spatiotemporal characteristics of regional latent dynamics:

\begin{equation}\label{eq:11}
\mathbf{H}^{(\ell)}=\operatorname{BiMamba}^{(\ell)}_{\mathrm{sp}}\!\left(\operatorname{Mamba}^{(\ell)}_{\mathrm{time}}\!\left(\mathbf{H}^{(\ell-1)}\right)\right)
\end{equation}

where \ensuremath{\mathbf{H}^{(\ell-1)}} and \ensuremath{\mathbf{H}^{(\ell)}} denote the input and output features of the \ensuremath{\ell} block, respectively.  \ensuremath{\operatorname{Mamba}^{(\ell)}_{\mathrm{time}}} represents temporal causal state-space scanning, while \ensuremath{\operatorname{BiMamba}^{(\ell)}_{\mathrm{sp}}} denotes bidirectional spatial state-space scanning and feature fusion.

After multiple spatiotemporal Mamba blocks, a linear readout layer maps the resulting hidden features to regional latent variables, producing the latent trajectories over the entire prediction horizon. The predicted latent variables are then passed through the shared coordinate-conditioned decoder to reconstruct the high-dimensional nodal response fields:

\begin{equation}\label{eq:12}
\widehat{\mathbf{Z}}_{0:T}=f_{\mathrm{out}}\!\left(\mathbf{H}^{(L)}\right),\quad \widehat{\mathbf{Y}}_t=\mathcal{D}_{\psi}\!\left(\widehat{\mathbf{Z}}_t;\mathbf{X},\mathcal{P}\right)
\end{equation}

where L is the number of spatiotemporal Mamba blocks,  \ensuremath{f_{\mathrm{out}}} denotes the linear readout mapping,  \ensuremath{\widehat{\mathbf{Z}}_{0:T}} represents the predicted regional latent trajectories, and \ensuremath{\widehat{\mathbf{Y}}_t} denotes the reconstructed nodal response field.

During the training of the spatiotemporal dynamics model, reference latent trajectories generated by the pretrained spatial encoder from ground-truth response fields are used as supervision targets. The optimization objective minimizes the discrepancy between the predicted and reference latent variables:

\begin{equation}\label{eq:13}
\mathcal{L}_{\mathrm{dyn}}=\frac{1}{T+1}\sum_{t=0}^{T}\left\lVert\widehat{\mathbf{Z}}_t-\mathbf{Z}_t\right\rVert_F^2
\end{equation}

where  \ensuremath{\mathbf{Z}_t} denotes the reference latent variables obtained from the pretrained spatial encoder,  \ensuremath{\widehat{\mathbf{Z}}_t} represents the predicted latent variables, and  \ensuremath{\lVert\cdot\rVert_F} denotes the Frobenius norm.

After Stage 2, the spatial encoder remains frozen, while the spatiotemporal dynamics model and shared decoder are jointly fine-tuned using physical response-field prediction errors. This final stage aims to reduce the mismatch between latent-state prediction and full-field reconstruction.

\section{Experimental setup}\label{sec:3}

\subsection{Datasets}\label{sec:3.1}

To evaluate the predictive capability of TPLD across different structural dynamics problems, three finite element datasets are considered: a parametric stiffened panel (SteelPanel), a partitioned pressure impact plate (PressurePlate), and a Toyota Yaris full-vehicle crash model. These datasets cover variations in both geometry and loading conditions, spatially distributed loading on a fixed structure, and large-scale structural dynamics under different initial impact velocities, respectively. The main characteristics of the datasets are summarized in Table \hyperref[tab:1]{1}.

The SteelPanel dataset is generated using Abaqus/Explicit finite element simulations. A total of 256 geometric configurations are constructed by varying the dimensions of the panel and its longitudinal and transverse stiffeners, with eight transient pressure loading cases applied to each configuration. The geometric and loading parameters are primarily sampled using scrambled Sobol sequences, with their ranges provided in Appendix A. The structure is discretized using four-node S4 shell elements with an elastoplastic material model. The applied pressure follows a symmetric triangular pulse defined by its peak magnitude and duration. To capture the rapid dynamic response during the initial loading stage, a nonuniform temporal sampling strategy is adopted, with output intervals of 0.01 ms and 0.1 ms over 0–0.7 ms and 0.7–10.7 ms, respectively. The three nodal displacement components (U1, U2, and U3) are selected as prediction targets. Regional partitioning is performed based on FE mesh topology and the structural characteristics of the stiffeners to construct a regional latent space with explicit structural correspondence.

The PressurePlate dataset is generated using Abaqus/Explicit simulations of a cross-stiffened plate with fixed geometry. The plate measures 1200 mm \ensuremath{\times} 1200 mm and is divided into nine independently loaded regions. Different loading cases are constructed by varying the pressure histories applied to these regions, representing dynamic responses under spatially nonuniform loading conditions. The FE model employs S4R shell elements and B31 beam elements, with all four plate edges fully constrained. The dataset covers a response duration of 0–10 ms, and the prediction targets include the three displacement components (U1, U2, and U3) and equivalent plastic strain (PEEQ). Based on the loading partitions and structural characteristics of the stiffeners, a multiscale regional latent space is constructed at three levels—coarse, medium, and fine—to represent dynamic response features at different spatial scales.

The Toyota Yaris full-vehicle crash dataset is generated from frontal impact simulations using LS-DYNA. A fixed full-vehicle FE model is employed, and different simulation cases are generated by varying the initial impact velocity between 40 and 60 km/h. All cases involve full-width frontal impacts and share identical mesh topology and node ordering. The response data span 0–150 ms, with a uniform sampling interval of 5 ms. The prediction targets are the three displacement components (U1, U2, and U3). Structural regions are organized according to vehicle components, mesh topology, and material properties, while connected mesh blocks are used for local feature extraction to accommodate the complexity of the full-vehicle structure. This dataset is primarily used to evaluate the capability of TPLD to predict full-field displacement histories in large-scale structural systems involving approximately 1.49 million nodes and complex nonlinear crash dynamics.

An additional body-in-white (BIW) automotive crash benchmark is introduced for comparison with existing surrogate models. Following the LS-DYNA-based benchmark considered by Akhare et al. (2026)(\hyperlink{ref:akhare2026}{Akhare et al., 2026}), the finite element model contains approximately 400,000 nodes and 380,000 elements. The simulations involve frontal impacts against a rigid barrier at an initial velocity of 56 km/h over a duration of 120 ms. Variations in structural configurations are introduced by perturbing the thicknesses of 33 front-end components within ±20\% of their nominal values, resulting in 150 simulation cases with 25 temporal frames per case. The three displacement components (U1, U2, and U3) are selected as prediction targets. This benchmark is used to assess TPLD against existing geometry-aware neural-operator models under a common crash prediction task.

\Needspace{145pt}
{\small\begin{longtable}{@{}>{\raggedright\arraybackslash}p{\dimexpr 0.25\linewidth-2.5\tabcolsep\relax} >{\raggedright\arraybackslash}p{\dimexpr 0.15\linewidth-1.5\tabcolsep\relax} >{\raggedright\arraybackslash}p{\dimexpr 0.15\linewidth-1.5\tabcolsep\relax} >{\raggedright\arraybackslash}p{\dimexpr 0.15\linewidth-1.5\tabcolsep\relax} >{\raggedright\arraybackslash}p{\dimexpr 0.15\linewidth-1.5\tabcolsep\relax} >{\raggedright\arraybackslash}p{\dimexpr 0.15\linewidth-1.5\tabcolsep\relax}@{}}
\caption{Summary of the structural dynamics datasets.}\label{tab:1}\\
\toprule
Dataset & Cases & Nodes & Frames & tokens & dim \\
\midrule\endfirsthead
\toprule
Dataset & Cases & Nodes & Frames & tokens & dim \\
\midrule\endhead\bottomrule\endlastfoot
SteelPanel & 2048 & 8k-10k & 171 & 41 & 128 \\
PressurePlate & 100 & 10k & 101 & 178 & 128 \\
BIW & 150 & 400k & 25 & 462 & 128 \\
Toyota Yaris & 116 & 1.5M & 31 & 5236 & 288 \\
\end{longtable}}

\subsection{Experimental Configuration}\label{sec:3.2}

To evaluate the predictive performance of TPLD under different structural and loading conditions, each dataset is divided into training, validation, and test sets at an approximate ratio of 8:1:1. For the SteelPanel dataset, a geometry-based group splitting strategy is adopted because each geometric configuration corresponds to eight loading cases. Specifically, the 256 geometric configurations are divided into 204, 26, and 26 configurations for training, validation, and testing, corresponding to 1,632, 208, and 208 simulation cases, respectively. All loading cases associated with the same geometric configuration are assigned to the same subset, ensuring no overlap in geometric configurations across the three subsets. This strategy enables the evaluation of model generalization to unseen geometries. The other datasets are split by simulation case, with all temporal frames from each case retained within the same subset.

The proposed model is implemented in PyTorch and trained on an NVIDIA A100 GPU. For the SteelPanel experiments, the spatial encoder consists of four MeshGraphNet layers, with a regional latent dimension of 128. Inter-region information exchange is performed using two attention layers, while the spatiotemporal dynamics model comprises four spatiotemporal Mamba blocks. The main training hyperparameters for each stage are summarized in Table \hyperref[tab:2]{2}. The training configurations for the other datasets are adjusted according to their structural scales and response characteristics.

The full-field time-history prediction performance of TPLD is evaluated on three structural dynamics datasets. In addition, a comparative experiment is conducted on the BIW crash benchmark to assess the predictive accuracy of TPLD against existing GeoTransolver variants under consistent experimental settings and evaluation protocols. Two types of ablation studies are conducted to investigate the contributions of the proposed mechanisms. Specifically, the SteelPanel dataset is used to examine the influence of different regional latent representations on prediction performance, while the PressurePlate dataset is used to assess the contribution of auxiliary one-step prediction to latent-space dynamical predictability. The corresponding results and discussions are presented in Section \hyperref[sec:4]{4}.

\Needspace{213pt}
{\small\begin{longtable}{@{}>{\raggedright\arraybackslash}p{\dimexpr 0.32\linewidth-0.64\tabcolsep\relax} >{\raggedright\arraybackslash}p{\dimexpr 0.68\linewidth-1.36\tabcolsep\relax}@{}}
\caption{Model architecture and training hyperparameters.}\label{tab:2}\\
\toprule
Component & Configuration \\
\midrule\endfirsthead
\toprule
Component & Configuration \\
\midrule\endhead\bottomrule\endlastfoot
Spatial encoder & 4 MGN layers \\
Regional representation & 128-dimensional latent variables \\
Region interaction & 2 attention layers \\
Spatiotemporal dynamics & 4 Mamba blocks \\
Stage 1 & 150 epochs, batch size 4, LR = 1 \ensuremath{\times} 10\textsuperscript{-}\textsuperscript{4} \\
Stage 2 & 600 epochs, batch size 1, LR = 1 \ensuremath{\times} 10\textsuperscript{-}\textsuperscript{4} \\
Stage 3 & 50 epochs, batch size 1, LR = 5 \ensuremath{\times} 10\textsuperscript{-}\textsuperscript{5} \\
Gradient accumulation & 4 steps \\
\end{longtable}}

\subsection{Evaluation Metrics}\label{sec:3.3}

To quantitatively evaluate the accuracy of TPLD in predicting full-field dynamic responses, three evaluation metrics are employed: relative L2 error, mean absolute error (MAE), and the coefficient of determination (R\textsuperscript{2}). Given the differences in mesh size, number of time frames, and response variables across the datasets, each metric is calculated separately for each response channel and then averaged over all test cases, ensuring equal statistical weight for each case.

The relative L2 error quantifies the overall prediction discrepancy relative to the magnitude of the ground-truth response field. MAE measures the average absolute difference between predicted and ground-truth nodal responses, while R\textsuperscript{2} evaluates how well the predictions capture variations in the ground-truth responses. These metrics are defined as follows:

\begin{equation}\label{eq:14}
E_{\mathrm{rel}}^{(c)} = \frac{1}{K}\sum_{k=1}^{K}\frac{\left\lVert\widehat{\mathbf{Y}}_k^{(c)}-\mathbf{Y}_k^{(c)}\right\rVert_F}{\left\lVert\mathbf{Y}_k^{(c)}\right\rVert_F+\varepsilon}
\end{equation}

\begin{equation}\label{eq:15}
\mathrm{MAE}^{(c)} = \frac{1}{K}\sum_{k=1}^{K}\frac{\left\lVert\widehat{\mathbf{Y}}_k^{(c)}-\mathbf{Y}_k^{(c)}\right\rVert_1}{N_kT_k}
\end{equation}

\begin{equation}\label{eq:16}
R_c^2 = \frac{1}{K}\sum_{k=1}^{K}\left(1-\frac{\left\lVert\widehat{\mathbf{Y}}_k^{(c)}-\mathbf{Y}_k^{(c)}\right\rVert_F^2}{\left\lVert\mathbf{Y}_k^{(c)}-\overline{\mathbf{Y}}_k^{(c)}\right\rVert_F^2+\varepsilon}\right)
\end{equation}

where K denotes the number of test cases, and N\textsubscript{k} and T\textsubscript{k} represent the numbers of nodes and time frames in the k-th case, respectively. The symbols Y and \ensuremath{\widehat{Y}} denote the ground-truth and predicted response fields, respectively. The superscript c identifies the corresponding physical response channel, and \ensuremath{\overline{Y}} represents the mean ground-truth response of that channel for the given case.

\section{Results and discussion}\label{sec:4}

\subsection{Prediction Performance Across Different Structures}\label{sec:4.1}

\begin{figure}[H]\centering
\includegraphics[width=0.92\linewidth,height=0.58\textheight,keepaspectratio]{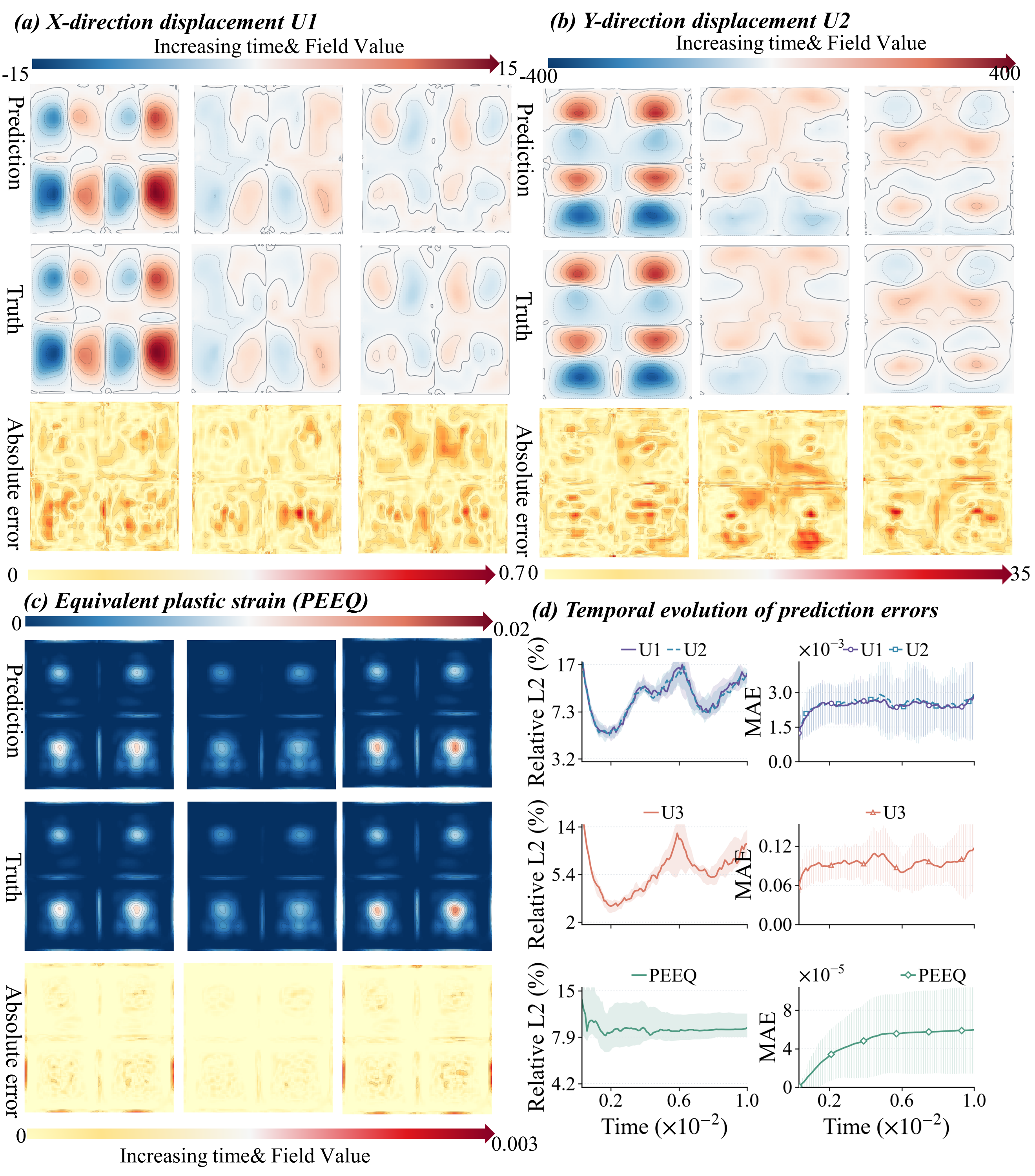}
\caption{Full-field response distributions and temporal evolution of prediction errors for the PressurePlate dataset.}\label{fig:4}
\end{figure}

To evaluate the predictive performance of TPLD across different structural and loading conditions, full-field time-history predictions are assessed on the PressurePlate, Toyota Yaris, and SteelPanel datasets. Figs. \hyperref[fig:4]{4}–6 compare representative predicted and finite element (FE) response fields and present the temporal evolution of prediction errors across all test cases. The corresponding quantitative results are summarized in Tables \hyperref[tab:4]{A2}–A4.

The PressurePlate dataset evaluates model performance under spatially nonuniform transient pressure loading. As shown in Fig. \hyperref[fig:4]{4}(a)–(c), TPLD captures the principal spatial distributions and temporal evolution of displacement and equivalent plastic strain, including localized response concentrations. The relative L2 errors are 8.57\%, 8.60\%, 4.58\%, and 7.75\% for U1, U2, U3, and PEEQ, respectively, with R\textsuperscript{2} values of 0.99 for all channels. Fig. \hyperref[fig:4]{4}(d) shows that displacement errors fluctuate during loading, whereas the relative L2 error of PEEQ remains comparatively stable despite a gradual increase in MAE.

\begin{figure}[H]\centering
\includegraphics[width=0.92\linewidth,height=0.58\textheight,keepaspectratio]{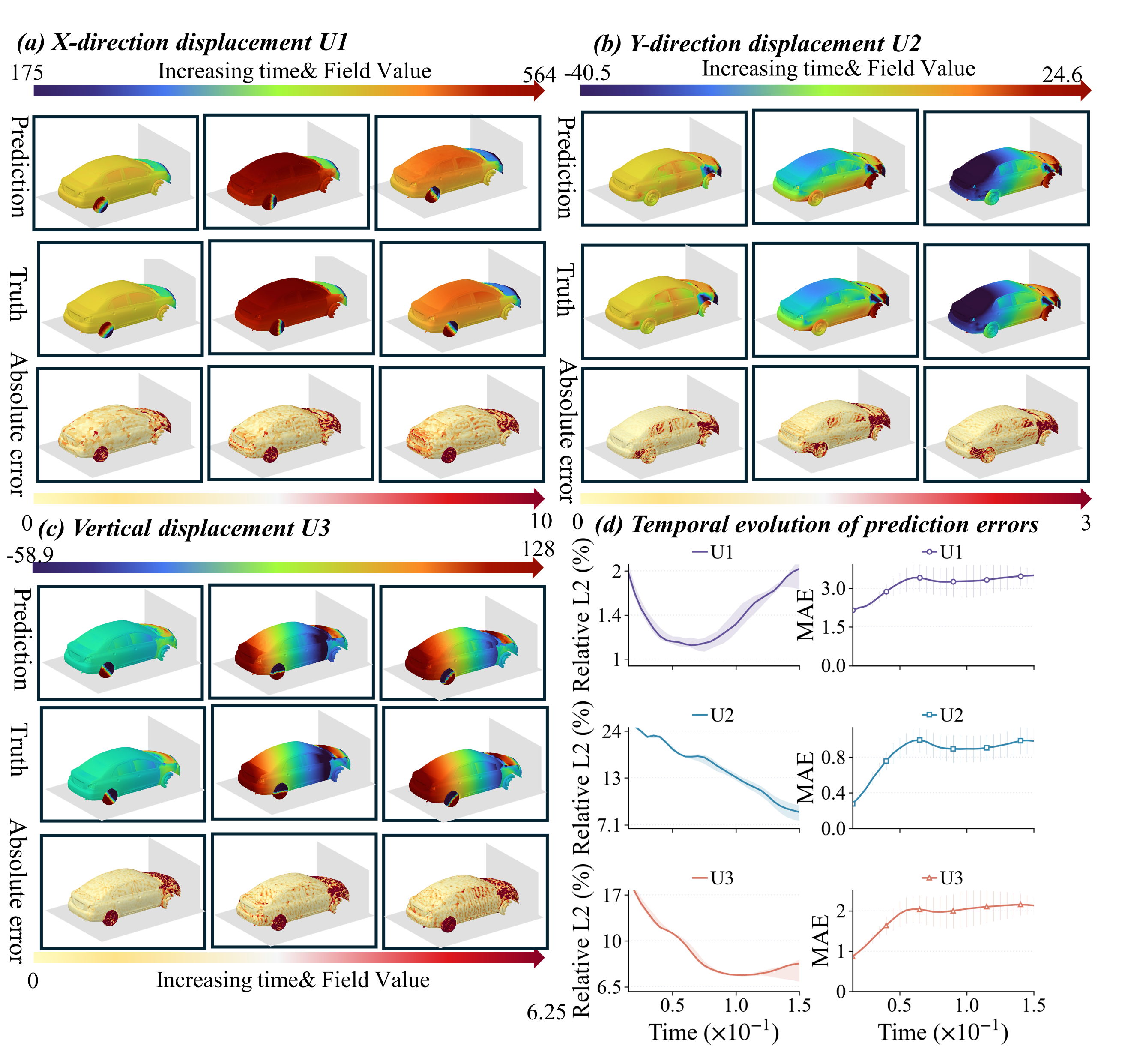}
\caption{Full-field displacement distributions and temporal evolution of prediction errors for the Toyota Yaris dataset.}\label{fig:5}
\end{figure}

The Toyota Yaris dataset examines full-field prediction for a large-scale vehicle model containing approximately 1.49 million nodes. As shown in Fig. \hyperref[fig:5]{5}(a)–(c), TPLD reproduces the principal deformation patterns during frontal impact, with relative L2 errors ranging from 1.41\% to 12.22\% and R\textsuperscript{2} values between 0.98 and 0.99. The temporal error curves in Fig. \hyperref[fig:5]{5}(d) show that relative errors generally decrease or remain low as the impact progresses, while MAEs increase during the initial stage before gradually stabilizing.

\begin{figure}[H]\centering
\includegraphics[width=0.92\linewidth,height=0.58\textheight,keepaspectratio]{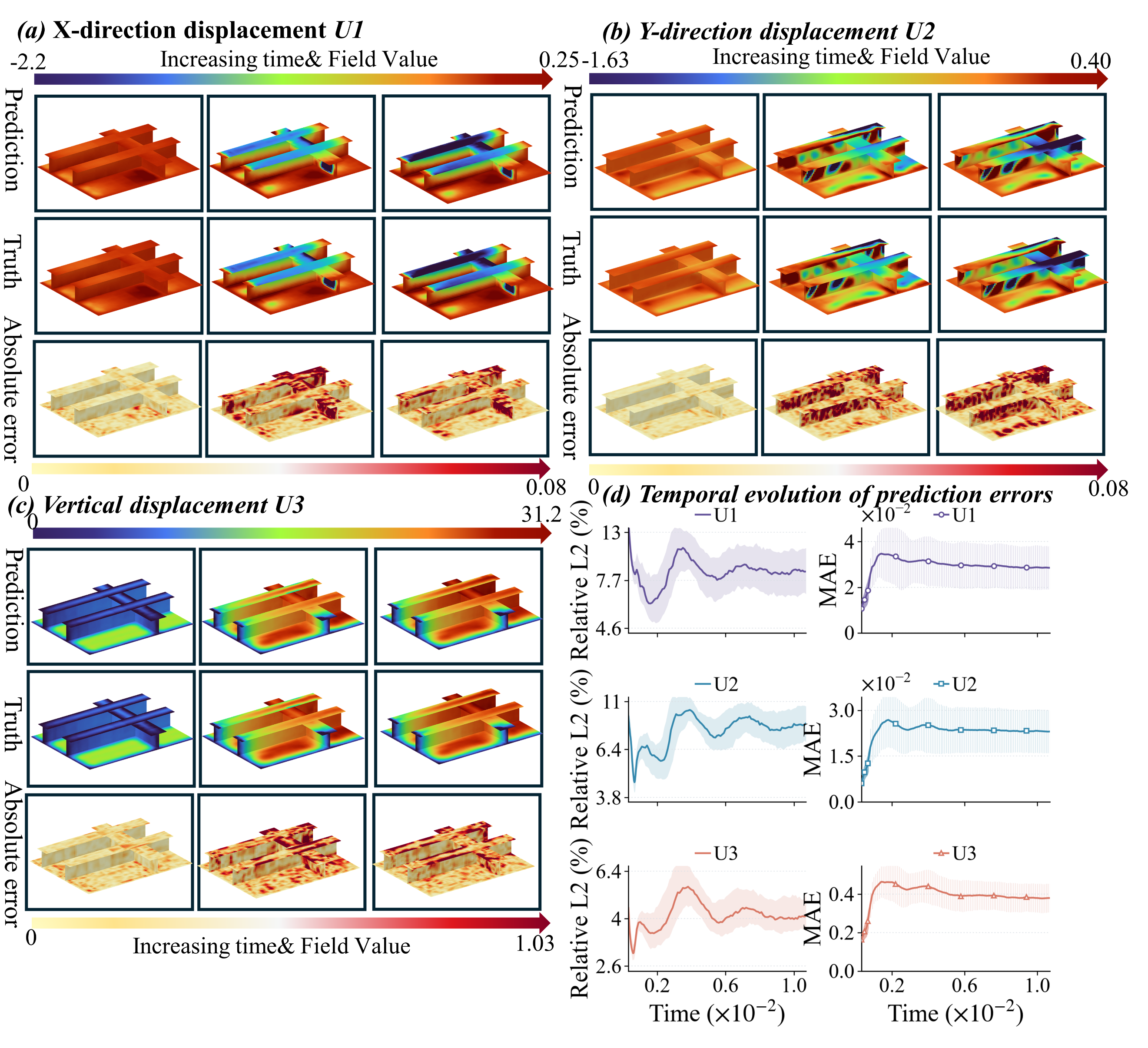}
\caption{Full-field displacement distributions and temporal evolution of prediction errors for the SteelPanel dataset.}\label{fig:6}
\end{figure}

The SteelPanel dataset further evaluates generalization to unseen geometric configurations under different transient pressure loads. Fig. \hyperref[fig:6]{6}(a)–(c) shows good agreement between the predicted and FE displacement fields in terms of the locations, magnitudes, and temporal evolution of the principal deformation regions. The relative L2 errors for U1, U2, and U3 are 8.21\%, 7.00\%, and 4.00\%, respectively, with R\textsuperscript{2} values of 0.99. As shown in Fig. \hyperref[fig:6]{6}(d), relative errors fluctuate initially before stabilizing, while MAEs increase during the early loading stage and subsequently remain approximately stable.

Taken together, the results demonstrate that TPLD can predict full-field displacement responses across structures with different mesh sizes, geometries, and loading conditions. The SteelPanel results demonstrate predictive performance on unseen geometric configurations, while the PressurePlate and Toyota Yaris results characterize model performance under spatially nonuniform loading and complex frontal impact conditions, respectively. Although prediction accuracy varies among displacement components and datasets, the model consistently captures the principal spatial deformation patterns and their temporal evolution.

\FloatBarrier
\subsection{Comparison with Existing Methods on the BIW Crash Benchmark}\label{sec:4.2}

To further assess the predictive performance of TPLD against existing surrogate models, a comparative experiment is conducted on the BIW automotive crash benchmark. The proposed method is compared with GeoTransolver and its FLARE-enhanced variant reported by Akhare et al. (2026). The comparison is performed using consistent dataset partitions and evaluation protocols, with relative L2 position error adopted as the quantitative metric.

Unlike the displacement-based relative L2 error defined in Section \hyperref[sec:3.3]{3.3}, the relative L2 position error evaluates the discrepancy between predicted and reference nodal positions, which are obtained by adding the corresponding displacement fields to the initial nodal coordinates. The position error is calculated for each test case and time step and subsequently averaged over the evaluation set:

\begin{equation}\label{eq:17}
P_{k,t}=X_{k,0}+U_{k,t},\qquad \widehat{P}_{k,t}=X_{k,0}+\widehat{U}_{k,t}
\end{equation}

\begin{equation}\label{eq:18}
E_{\mathrm{pos}}=\frac{100}{K}\sum_{k=1}^{K}\frac{1}{T_k}\sum_{t=1}^{T_k}\frac{\left\lVert\widehat{P}_{k,t}-P_{k,t}\right\rVert_F}{\left\lVert P_{k,t}\right\rVert_F+\varepsilon}
\end{equation}

where  \ensuremath{X_{k,0}}  denotes the initial nodal coordinates,  \ensuremath{U_{k,t}} and  \ensuremath{\widehat{U}_{k,t}}  are the reference and predicted displacements, and \ensuremath{P_{k,t}} and \ensuremath{\widehat{P}_{k,t}} are the corresponding nodal positions.  \ensuremath{K} and  \ensuremath{T_k} denote the numbers of test cases and evaluated time steps, respectively,  \ensuremath{\lVert\cdot\rVert_F} is the Frobenius norm, and \ensuremath{\varepsilon} is a small constant preventing division by zero.

\begin{figure}[!htb]\centering
\includegraphics[width=0.96\linewidth,height=0.63\textheight,keepaspectratio]{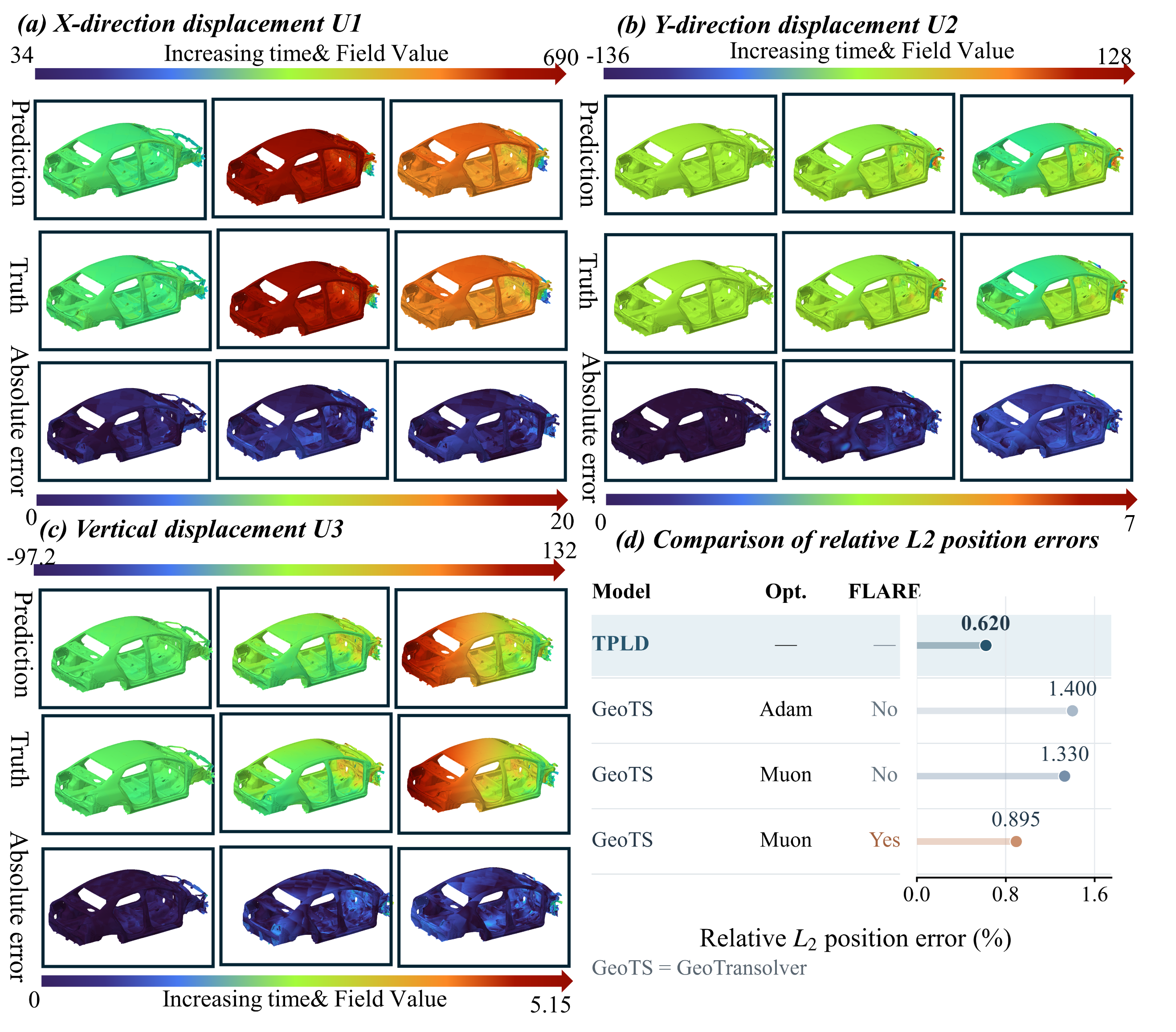}
\caption{Full-field displacement predictions and comparison with GeoTransolver on the BIW crash benchmark.}\label{fig:7}
\end{figure}

Fig. \hyperref[fig:7]{7}(a)–(c) presents the predicted displacement fields, finite element reference fields, and corresponding absolute errors for the three displacement components at 20, 60, and 120 ms. As the collision progresses, the structural deformation develops progressively, exhibiting distinct spatial distributions in different displacement directions. TPLD captures the principal deformation patterns and their temporal evolution, with the predicted fields showing good agreement with the FE reference results. Although localized discrepancies remain, the overall spatial distributions and deformation trends are well reproduced throughout the selected time steps.

The quantitative comparison is presented in Fig. \hyperref[fig:7]{7}(d). TPLD achieves a relative L2 position error of 0.620\%, compared with 1.400\% and 1.330\% for GeoTransolver trained with Adam and Muon, respectively. The FLARE-enhanced GeoTransolver trained with Muon achieves an error of 0.895\%. Among the evaluated methods, TPLD yields the lowest position error, representing a reduction of approximately 30.7\% relative to the best-performing GeoTransolver configuration. These results demonstrate the competitive accuracy of the proposed framework in predicting full-field nodal trajectories under nonlinear automotive impact conditions. Combined with the ablation results presented in Section \hyperref[sec:4.3]{4.3}, this comparison supports the effectiveness of the proposed regional latent dynamics framework for transient structural response prediction.

\FloatBarrier
\subsection{Ablation Studies}\label{sec:4.3}

Two ablation experiments are conducted to evaluate the contributions of topology-guided regional representations and auxiliary one-step prediction. On SteelPanel, TPLD is compared with global latent representation (Global latent) and learnable soft-slice representation (Soft slice). Global latent aggregates nodal features through attention-based pooling, whereas Soft slice employs learned node-to-slice assignments following Transolver(\hyperlink{ref:wu2024}{Wu H et al., 2024}). All variants use 41 latent variables of 128 dimensions and share the same dynamics architecture and training protocol. On PressurePlate, auxiliary one-step prediction is removed (Without one-step), while the remaining configurations are unchanged.

To compare prediction performance across time, a dimensionless composite score is constructed from relative L2 error, MAE normalized by the standard deviation of the reference responses, and the R\textsuperscript{2}-based error term. The three components are equally weighted, and higher scores indicate better overall prediction performance. The composite score is defined as

\begin{equation}\label{eq:19}
S=\frac{100}{1+\frac{1}{3}\left(e_{L2}+e_{\mathrm{MAE}}+e_{R2}\right)}
\end{equation}

\begin{equation}\label{eq:20}
e_{L2}=\frac{\left\lVert\widehat{\mathbf{y}}-\mathbf{y}\right\rVert_2}{\left\lVert\mathbf{y}\right\rVert_2},\quad e_{\mathrm{MAE}}=\frac{\mathrm{MAE}}{\sigma_y},\quad e_{R2}=\sqrt{1-R^2}
\end{equation}

\begin{figure}[!htb]
    \centering
    \includegraphics[
        width=0.95\linewidth,
        height=0.52\textheight,
        keepaspectratio
    ]{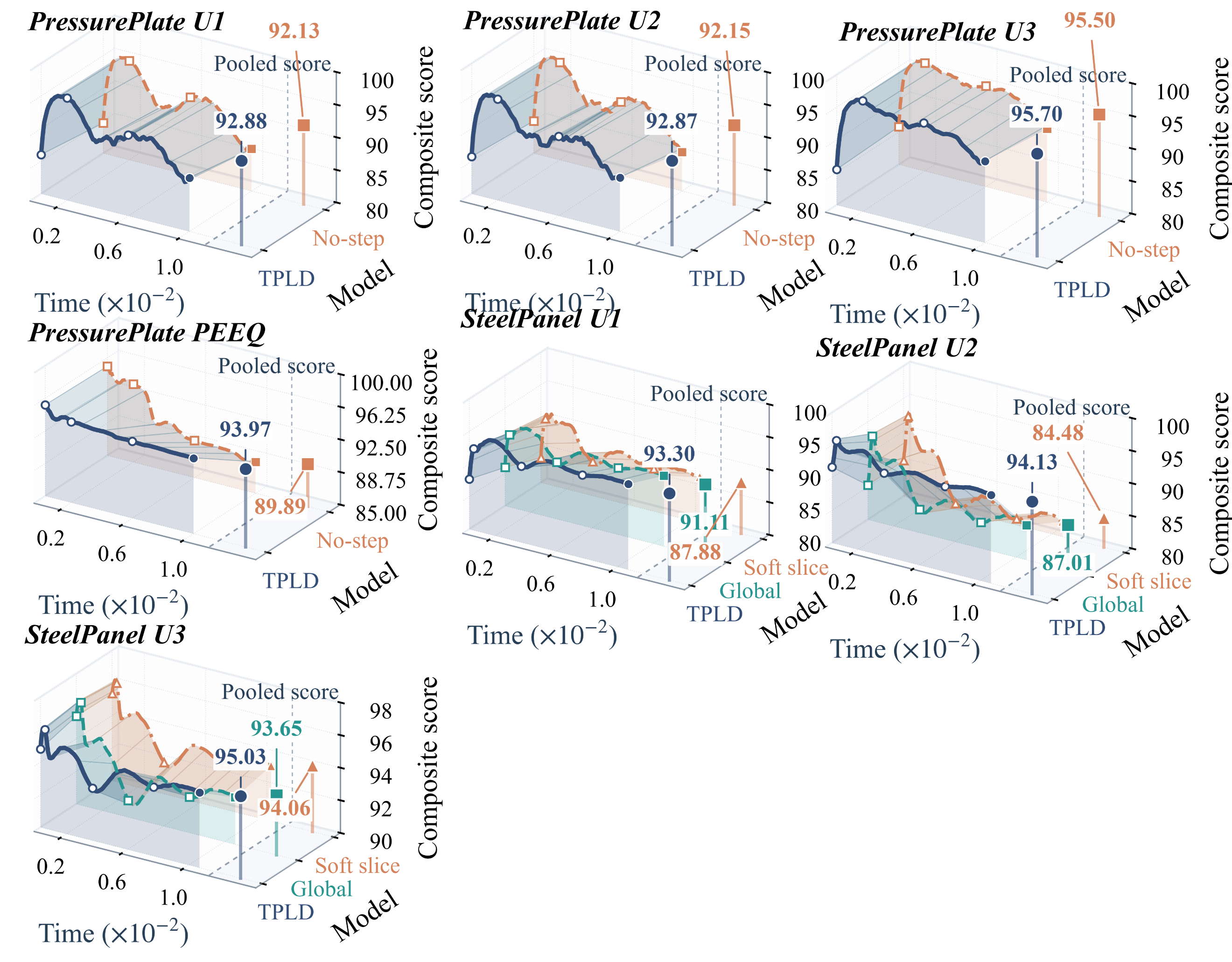}
    \caption{Temporal evolution of composite prediction scores
    in the ablation studies on the PressurePlate and
    SteelPanel datasets.}
    \label{fig:8}
\end{figure}

Here,  \ensuremath{\sigma_y} denotes the standard deviation of the ground-truth responses over the corresponding evaluation domain. The three error terms are equally weighted, and a higher composite score indicates better overall prediction performance. Fig. \hyperref[fig:8]{8} presents the temporal evolution and aggregated values of the composite scores, while the conventional evaluation metrics are reported in Tables \hyperref[tab:7]{A5} and A6.

For the regional representation ablation on SteelPanel, TPLD achieves the highest aggregated composite scores for all three displacement components (Fig. \hyperref[fig:8]{8}). The improvement is particularly evident for U2, where TPLD obtains a score of 94.13, compared with 87.01 for Global latent and 84.48 for Soft slice. Consistently, the corresponding relative L2 error is reduced to 7.00\%, compared with 16.97\% and 16.30\%, respectively (Table \hyperref[tab:7]{A5}). These results indicate that topology-guided regional organization improves prediction performance under matched latent capacities and training configurations, suggesting the value of preserving explicit structural correspondence in latent representations.

For the auxiliary one-step prediction ablation on PressurePlate, the complete TPLD model achieves higher aggregated composite scores across all four response channels (Fig. \hyperref[fig:8]{8}). The most pronounced improvement occurs for PEEQ, with the score increasing from 89.89 to 93.97. The corresponding relative L2 error decreases from 14.10\% to 7.75\% (Table \hyperref[tab:8]{A6}), while improvements in the displacement components are comparatively modest. These findings suggest that auxiliary one-step prediction enhances the dynamical predictability of learned representations, particularly for responses involving history-dependent plastic strain accumulation.

\FloatBarrier

\subsection{Discussion}\label{sec:4.4}

Across the evaluated datasets, TPLD demonstrates the capability to predict full-field response histories for structures ranging from parametrically varied stiffened panels to a large-scale vehicle model. The SteelPanel experiments evaluate generalization to unseen geometric configurations, whereas PressurePlate and Toyota Yaris examine different loading conditions on fixed meshes. These evaluations demonstrate the applicability of the framework across different structural complexities, mesh sizes, and loading conditions.

The BIW crash benchmark provides a direct comparison with existing geometry-aware surrogate models. TPLD achieves a relative L2 position error of 0.620\%, compared with 0.895\% for the best-performing FLARE-enhanced GeoTransolver configuration, representing a 30.7\% reduction. This result demonstrates the competitive predictive accuracy of TPLD on a common automotive crash benchmark. Unlike GeoTransolver's learnable physical-slice representations, TPLD incorporates physically defined regional latent variables and load-conditioned spatiotemporal dynamics. However, the comparison evaluates the complete frameworks rather than isolating the contribution of individual architectural differences.

The SteelPanel ablation further indicates that latent-space organization influences prediction accuracy under matched latent capacities and downstream configurations. By assigning latent variables to physically defined regions and constraining their interactions through mesh topology, TPLD may better preserve localized response characteristics than global pooling or learnable soft slicing. This suggests that incorporating structural priors into latent-space organization can benefit subsequent dynamical prediction.

The PressurePlate ablation highlights the complementary role of dynamical supervision. Reconstruction-based learning constrains the representation of response fields at individual time steps but does not explicitly constrain their evolution. Auxiliary one-step prediction introduces supervision of latent-state transitions, with particularly pronounced improvements in PEEQ prediction. This finding is consistent with the history-dependent nature of plastic strain accumulation and suggests that dynamical supervision can improve the suitability of learned representations for time-history prediction.

Temporal error trends also require careful interpretation. Because relative L2 error is normalized by the magnitude of the reference response, small deformations during early loading may produce relatively large relative errors. As deformation develops, relative errors can decrease even when MAE increases. The increase in PEEQ MAE alongside relatively stable relative errors further highlights the importance of considering both relative and absolute errors when evaluating transient structural responses.

\section{Conclusions}\label{sec:5}

This study proposes a topology-prior-guided regional latent dynamics (TPLD) surrogate model for full-field time-history prediction of transient structural responses. By combining physically structured regional representations with dynamics-aware latent learning and load-conditioned spatiotemporal modeling, the framework addresses the need to preserve local response characteristics while maintaining latent-space dynamical predictability. The main findings are summarized as follows:

\noindent\textbullet\quad Topology-prior-guided regional representations improve prediction accuracy under matched experimental settings. On SteelPanel, TPLD achieves lower relative L2 errors than global latent and learnable soft-slice representations across all three displacement components. For U2, the error decreases to 7.00\%, compared with 16.97\% and 16.30\%, respectively. These results demonstrate the benefits of incorporating engineering priors into regional latent-space organization.

\noindent\textbullet\quad TPLD enables full-field time-history prediction across different structural geometries, mesh sizes, and loading conditions. The relative L2 errors for the three displacement components range from 4.00\% to 8.21\% on SteelPanel and from 1.41\% to 12.22\% on Toyota Yaris, which contains approximately 1.49 million nodes. On the BIW automotive crash benchmark, TPLD achieves a relative L2 position error of 0.620\%, compared with 0.895\% for the best-performing FLARE-enhanced GeoTransolver configuration, representing a 30.7\% reduction. These results demonstrate the predictive capability of TPLD across different structural geometries, mesh sizes, and transient loading conditions, as well as its superior accuracy over the evaluated baseline methods on the BIW benchmark.

\noindent\textbullet\quad Auxiliary one-step prediction improves latent-space dynamical predictability through additional state-transition supervision. The PressurePlate ablation shows improvements across all evaluated response channels, with the most pronounced benefit observed for equivalent plastic strain (PEEQ). Removing auxiliary one-step prediction increases the PEEQ relative L2 error from 7.75\% to 14.10\%. This finding highlights the value of dynamical supervision for predicting history-dependent structural responses.

Several limitations remain. Regional partitioning currently relies on prescribed engineering priors, and model performance has been evaluated primarily within predefined structural and loading parameter ranges using numerical FE datasets. Future work will investigate adaptive region partitioning, generalization to broader structural and loading conditions, and validation using experimental measurements to further assess the practical applicability of TPLD.

\clearpage\section*{Appendix. A}\phantomsection\label{app:A}\setcounter{table}{0}\renewcommand{\thetable}{A\arabic{table}}\renewcommand{\theHtable}{A\arabic{table}}

\Needspace{264pt}
{\small\begin{longtable}{@{}>{\raggedright\arraybackslash}p{\dimexpr 0.14\linewidth-0.56\tabcolsep\relax} >{\raggedright\arraybackslash}p{\dimexpr 0.58\linewidth-2.32\tabcolsep\relax} >{\raggedright\arraybackslash}p{\dimexpr 0.28\linewidth-1.12\tabcolsep\relax}@{}}
\caption{Geometric and loading parameters of the stiffened panel dataset.}\label{tab:3}\\
\toprule
Parameter & Description & Range \\
\midrule\endfirsthead
\toprule
Parameter & Description & Range \\
\midrule\endhead\bottomrule\endlastfoot
\ensuremath{L} & Panel length (mm) & 1351–1650 \\
\ensuremath{W} & Panel width (mm) & 1080–1320 \\
\ensuremath{t} & Panel thickness (mm) & 1.6–2.4 \\
\ensuremath{t_L} & Longitudinal stiffener thickness (mm) & 1.2–1.8 \\
\ensuremath{h_L} & Longitudinal stiffener height (mm) & 90–130 \\
\ensuremath{l_L} & Longitudinal stiffener flange width (mm) & 85–125 \\
\ensuremath{t_W} & Transverse stiffener thickness (mm) & 1.2–1.8 \\
\ensuremath{h_W} & Transverse stiffener height (mm) & 90–130 \\
\ensuremath{l_W} & Transverse stiffener flange width (mm) & 85–125 \\
\ensuremath{p_{peak}} & Peak pressure (MPa) & 4.0–5.99 \\
\ensuremath{t_{end}} & Pulse duration (ms) & 0.24–0.36 \\
\end{longtable}}

\Needspace{128pt}
{\small\begin{longtable}{@{}>{\raggedright\arraybackslash}p{\dimexpr 0.24\linewidth-1.44\tabcolsep\relax} >{\raggedright\arraybackslash}p{\dimexpr 0.2\linewidth-1.2000000000000002\tabcolsep\relax} >{\raggedright\arraybackslash}p{\dimexpr 0.25\linewidth-1.5\tabcolsep\relax} >{\raggedright\arraybackslash}p{\dimexpr 0.31\linewidth-1.8599999999999999\tabcolsep\relax}@{}}
\caption{Toyota Yaris.}\label{tab:4}\\
\toprule
Response & R\textsuperscript{2} & MAE & Relative L2 (\%) \\
\midrule\endfirsthead
\toprule
Response & R\textsuperscript{2} & MAE & Relative L2 (\%) \\
\midrule\endhead\bottomrule\endlastfoot
U1 & 0.99 & 3.09 & 1.41 \\
U2 & 0.98 & 0.80 & 12.22 \\
U3 & 0.99 & 1.81 & 8.19 \\
\end{longtable}}

\Needspace{128pt}
{\small\begin{longtable}{@{}>{\raggedright\arraybackslash}p{\dimexpr 0.24\linewidth-1.44\tabcolsep\relax} >{\raggedright\arraybackslash}p{\dimexpr 0.2\linewidth-1.2000000000000002\tabcolsep\relax} >{\raggedright\arraybackslash}p{\dimexpr 0.25\linewidth-1.5\tabcolsep\relax} >{\raggedright\arraybackslash}p{\dimexpr 0.31\linewidth-1.8599999999999999\tabcolsep\relax}@{}}
\caption{Stiffened steel panel.}\label{tab:5}\\
\toprule
Response & R\textsuperscript{2} & MAE & Relative L2 (\%) \\
\midrule\endfirsthead
\toprule
Response & R\textsuperscript{2} & MAE & Relative L2 (\%) \\
\midrule\endhead\bottomrule\endlastfoot
U1 & 0.99 & 0.02 & 8.21 \\
U2 & 0.99 & 0.02 & 7.00 \\
U3 & 0.99 & 0.31 & 4.00 \\
\end{longtable}}

\Needspace{145pt}
{\small\begin{longtable}{@{}>{\raggedright\arraybackslash}p{\dimexpr 0.24\linewidth-1.44\tabcolsep\relax} >{\raggedright\arraybackslash}p{\dimexpr 0.2\linewidth-1.2000000000000002\tabcolsep\relax} >{\raggedright\arraybackslash}p{\dimexpr 0.25\linewidth-1.5\tabcolsep\relax} >{\raggedright\arraybackslash}p{\dimexpr 0.31\linewidth-1.8599999999999999\tabcolsep\relax}@{}}
\caption{Partitioned pressure impact plate}\label{tab:6}\\
\toprule
Response & R\textsuperscript{2} & MAE & Relative L2 (\%) \\
\midrule\endfirsthead
\toprule
Response & R\textsuperscript{2} & MAE & Relative L2 (\%) \\
\midrule\endhead\bottomrule\endlastfoot
U1 & 0.99 & 2.44 \ensuremath{\times} 10\textsuperscript{-3} & 8.57 \\
U2 & 0.99 & 2.49 \ensuremath{\times} 10\textsuperscript{-3} & 8.60 \\
U3 & 0.99 & 9.37 \ensuremath{\times} 10\textsuperscript{-2} & 4.58 \\
PEEQ & 0.99 & 4.65 \ensuremath{\times} 10\textsuperscript{-5} & 7.75 \\
\end{longtable}}

\Needspace{230pt}
{\small\begin{longtable}{@{}>{\raggedright\arraybackslash}p{\dimexpr 0.23\linewidth-1.84\tabcolsep\relax} >{\raggedright\arraybackslash}p{\dimexpr 0.15\linewidth-1.2\tabcolsep\relax} >{\raggedright\arraybackslash}p{\dimexpr 0.14\linewidth-1.12\tabcolsep\relax} >{\raggedright\arraybackslash}p{\dimexpr 0.22\linewidth-1.76\tabcolsep\relax} >{\raggedright\arraybackslash}p{\dimexpr 0.26\linewidth-2.08\tabcolsep\relax}@{}}
\caption{Ablation study of regional latent representations on the SteelPanel dataset.}\label{tab:7}\\
\toprule
Model & Response & R\textsuperscript{2} & MAE & Relative L2 (\%) \\
\midrule\endfirsthead
\toprule
Model & Response & R\textsuperscript{2} & MAE & Relative L2 (\%) \\
\midrule\endhead\bottomrule\endlastfoot
TPLD & U1 & 0.99 & 0.02 & 8.21 \\
 & U2 & 0.99 & 0.02 & 7.00 \\
 & U3 & 0.99 & 0.31 & 4.00 \\
Global latent & U1 & 0.99 & 3.44 \ensuremath{\times} 10\textsuperscript{-2} & 11.21 \\
 & U2 & 0.97 & 4.52 \ensuremath{\times} 10\textsuperscript{-2} & 16.97 \\
 & U3 & 0.99 & 0.44 & 5.23 \\
Soft slice & U1 & 0.99 & 3.32 \ensuremath{\times} 10\textsuperscript{-2} & 10.02 \\
 & U2 & 0.97 & 4.38 \ensuremath{\times} 10\textsuperscript{-2} & 16.30 \\
 & U3 & 0.99 & 0.43 & 4.82 \\
\end{longtable}}

\Needspace{213pt}
{\small\begin{longtable}{@{}>{\raggedright\arraybackslash}p{\dimexpr 0.23\linewidth-1.84\tabcolsep\relax} >{\raggedright\arraybackslash}p{\dimexpr 0.15\linewidth-1.2\tabcolsep\relax} >{\raggedright\arraybackslash}p{\dimexpr 0.14\linewidth-1.12\tabcolsep\relax} >{\raggedright\arraybackslash}p{\dimexpr 0.22\linewidth-1.76\tabcolsep\relax} >{\raggedright\arraybackslash}p{\dimexpr 0.26\linewidth-2.08\tabcolsep\relax}@{}}
\caption{Ablation study of auxiliary one-step prediction on the PressurePlate dataset.}\label{tab:8}\\
\toprule
Model & Response & R\textsuperscript{2} & MAE & Relative L2 (\%) \\
\midrule\endfirsthead
\toprule
Model & Response & R\textsuperscript{2} & MAE & Relative L2 (\%) \\
\midrule\endhead\bottomrule\endlastfoot
TPLD & U1 & 0.99 & 2.44 \ensuremath{\times} 10\textsuperscript{-3} & 8.57 \\
 & U2 & 0.99 & 2.49 \ensuremath{\times} 10\textsuperscript{-3} & 8.60 \\
 & U3 & 0.99 & 9.37 \ensuremath{\times} 10\textsuperscript{-2} & 4.58 \\
 & PEEQ & 0.99 & 4.65 \ensuremath{\times} 10\textsuperscript{-5} & 7.75 \\
Without one-step & U1 & 0.99 & 2.72 \ensuremath{\times} 10\textsuperscript{-3} & 9.55 \\
 & U2 & 0.99 & 2.77 \ensuremath{\times} 10\textsuperscript{-3} & 9.53 \\
 & U3 & 0.99 & 9.76 \ensuremath{\times} 10\textsuperscript{-2} & 4.83 \\
 & PEEQ & 0.98 & 6.64 \ensuremath{\times} 10\textsuperscript{-5} & 14.10 \\
\end{longtable}}

\clearpage\section*{References}\begin{samepage}\noindent\hypertarget{ref:akhare2026}{}\hangindent=1.5em\hangafter=1 Akhare, D., Nabian, M.A., Adams, C., Chavare, S., Choudhry, S., 2026. High-Fidelity Industrial Crash Dynamics Prediction via Geometry-Aware Operator Learning with Memory-Efficient Low-Rank Attention. ArXiv abs/2605.27758.\par\end{samepage}\vspace{0.5em}

\begin{samepage}\noindent\hypertarget{ref:benner2015}{}\hangindent=1.5em\hangafter=1 Benner, P., Gugercin, S., Willcox, K., 2015. A Survey of Projection-Based Model Reduction Methods for Parametric Dynamical Systems. SIAM Rev. 57, 483–531. \url{https://doi.org/10.1137/130932715}\par\end{samepage}\vspace{0.5em}

\begin{samepage}\noindent\hypertarget{ref:brandstetter2022}{}\hangindent=1.5em\hangafter=1 Brandstetter J, Worrall D, Welling M, 2022. Message passing neural PDE solvers. Presented at the International Conference on Learning Representations, OpenReview.\par\end{samepage}\vspace{0.5em}

\begin{samepage}\noindent\hypertarget{ref:champion2019}{}\hangindent=1.5em\hangafter=1 Champion, K., Lusch, B., Kutz, J.N., Brunton, S.L., 2019. Data-driven discovery of coordinates and governing equations. Proc. Natl. Acad. Sci. 116, 22445–22451. \url{https://doi.org/10.1073/pnas.1906995116}\par\end{samepage}\vspace{0.5em}

\begin{samepage}\noindent\hypertarget{ref:dao2024}{}\hangindent=1.5em\hangafter=1 Dao, T., Gu, A., 2024. Transformers are SSMs: Generalized Models and Efficient Algorithms Through Structured State Space Duality. \url{https://doi.org/10.48550/ARXIV.2405.21060}\par\end{samepage}\vspace{0.5em}

\begin{samepage}\noindent\hypertarget{ref:du2026}{}\hangindent=1.5em\hangafter=1 Du, X., Wen, Z., Wang, HaiHua, Feng, Y., Xu, F., Shi, J., Wang, Hu, Li, G., 2026. A spatiotemporal discrete graph operator network for learning transient dynamics on irregular meshes. Comput. Struct. 330, 108341. \url{https://doi.org/10.1016/j.compstruc.2026.108341}\par\end{samepage}\vspace{0.5em}

\begin{samepage}\noindent\hypertarget{ref:fresca2021}{}\hangindent=1.5em\hangafter=1 Fresca, S., Dede’, L., Manzoni, A., 2021. A Comprehensive Deep Learning-Based Approach to Reduced Order Modeling of Nonlinear Time-Dependent Parametrized PDEs. J. Sci. Comput. 87. \url{https://doi.org/10.1007/s10915-021-01462-7}\par\end{samepage}\vspace{0.5em}

\begin{samepage}\noindent\hypertarget{ref:fresca2022}{}\hangindent=1.5em\hangafter=1 Fresca, S., Manzoni, A., 2022. POD-DL-ROM: Enhancing deep learning-based reduced order models for nonlinear parametrized PDEs by proper orthogonal decomposition. Comput. Methods Appl. Mech. Eng. 388, 114181. \url{https://doi.org/10.1016/j.cma.2021.114181}\par\end{samepage}\vspace{0.5em}

\begin{samepage}\noindent\hypertarget{ref:fries2022}{}\hangindent=1.5em\hangafter=1 Fries, W.D., He, X., Choi, Y., 2022. LaSDI: Parametric Latent Space Dynamics Identification. Comput. Methods Appl. Mech. Eng. 399, 115436. \url{https://doi.org/10.1016/j.cma.2022.115436}\par\end{samepage}\vspace{0.5em}

\begin{samepage}\noindent\hypertarget{ref:gu2023}{}\hangindent=1.5em\hangafter=1 Gu, A., Dao, T., 2023. Mamba: Linear-Time Sequence Modeling with Selective State Spaces. arXiv. \url{https://doi.org/10.48550/arXiv.2312.00752}\par\end{samepage}\vspace{0.5em}

\begin{samepage}\noindent\hypertarget{ref:gu2021}{}\hangindent=1.5em\hangafter=1 Gu A, Goel K, Ré C, 2021. Efficiently modeling long sequences with structured state spaces. arXiv 2111.00396.\par\end{samepage}\vspace{0.5em}

\begin{samepage}\noindent\hypertarget{ref:jaegle2022}{}\hangindent=1.5em\hangafter=1 Jaegle A, Borgeaud S, Alayrac JB, 2022. Perceiver IO: A general architecture for structured inputs \& outputs. Presented at the International Conference on Learning Representations, OpenReview.\par\end{samepage}\vspace{0.5em}

\begin{samepage}\noindent\hypertarget{ref:kneifl2025}{}\hangindent=1.5em\hangafter=1 Kneifl, J., Fehr, J., Brunton, S.L., Kutz, J.N., 2025. Multi-hierarchical surrogate learning for explicit structural dynamical systems using graph convolutional neural networks. Comput. Mech. 75, 1115–1135. \url{https://doi.org/10.1007/s00466-024-02553-6}\par\end{samepage}\vspace{0.5em}

\begin{samepage}\noindent\hypertarget{ref:lee2019}{}\hangindent=1.5em\hangafter=1 Lee J, Lee Y, Kim J, 2019. Set Transformer: A framework for attention-based permutation-invariant neural networks. Presented at the Proceedings of the 36th International Conference on Machine Learning, PMLR, pp. 3744–3753.\par\end{samepage}\vspace{0.5em}

\begin{samepage}\noindent\hypertarget{ref:lee2020}{}\hangindent=1.5em\hangafter=1 Lee, K., Carlberg, K.T., 2020. Model reduction of dynamical systems on nonlinear manifolds using deep convolutional autoencoders. J. Comput. Phys. 404, 108973. \url{https://doi.org/10.1016/j.jcp.2019.108973}\par\end{samepage}\vspace{0.5em}

\begin{samepage}\noindent\hypertarget{ref:li2023}{}\hangindent=1.5em\hangafter=1 Li, Q., Wang, Z., Li, L., Hao, H., Chen, W., Shao, Y., 2023. Machine learning prediction of structural dynamic responses using graph neural networks. Comput. Struct. 289, 107188. \url{https://doi.org/10.1016/j.compstruc.2023.107188}\par\end{samepage}\vspace{0.5em}

\begin{samepage}\noindent\hypertarget{ref:li2025}{}\hangindent=1.5em\hangafter=1 Li, Y., Wang, C., Hou, W., 2025. A universal surrogate modeling method based on heterogeneous graph neural network for nonlinear analysis. Comput. Methods Appl. Mech. Eng. 437, 117793. \url{https://doi.org/10.1016/j.cma.2025.117793}\par\end{samepage}\vspace{0.5em}

\begin{samepage}\noindent\hypertarget{ref:liz2023}{}\hangindent=1.5em\hangafter=1 Li Z, Kovachki N, Choy C, 2023. Geometry-informed neural operator for large-scale 3D PDEs. Presented at the Advances in Neural Information Processing Systems, NeurIPS.\par\end{samepage}\vspace{0.5em}

\begin{samepage}\noindent\hypertarget{ref:liu2024}{}\hangindent=1.5em\hangafter=1 Liu, Yue, Tian, Y., Zhao, Y., Yu, H., Xie, L., Wang, Y., Ye, Q., Jiao, J., Liu, Yunfan, 2024. VMamba: Visual State Space Model. \url{https://doi.org/10.48550/ARXIV.2401.10166}\par\end{samepage}\vspace{0.5em}

\begin{samepage}\noindent\hypertarget{ref:lusch2018}{}\hangindent=1.5em\hangafter=1 Lusch, B., Kutz, J.N., Brunton, S.L., 2018. Deep learning for universal linear embeddings of nonlinear dynamics. Nat. Commun. 9. \url{https://doi.org/10.1038/s41467-018-07210-0}\par\end{samepage}\vspace{0.5em}

\begin{samepage}\noindent\hypertarget{ref:nikolopoulos2022}{}\hangindent=1.5em\hangafter=1 Nikolopoulos, S., Kalogeris, I., Papadopoulos, V., 2022. Machine learning accelerated transient analysis of stochastic nonlinear structures. Eng. Struct. 257, 114020. \url{https://doi.org/10.1016/j.engstruct.2022.114020}\par\end{samepage}\vspace{0.5em}

\begin{samepage}\noindent\hypertarget{ref:perez2018}{}\hangindent=1.5em\hangafter=1 Perez E, Strub F, de Vries H, Dumoulin V, Courville A, 2018. FiLM: Visual reasoning with a general conditioning layer. Presented at the Proceedings of the AAAI Conference on Artificial Intelligence, AAAI Press, pp. 3942–3951.\par\end{samepage}\vspace{0.5em}

\begin{samepage}\noindent\hypertarget{ref:pfaff2021}{}\hangindent=1.5em\hangafter=1 Pfaff, T., Fortunato, M., Sanchez-Gonzalez, A., Battaglia, P.W., 2021. Learning Mesh-Based Simulation with Graph Networks. Presented at the International Conference on Learning Representations. \url{https://doi.org/10.48550/arXiv.2010.03409}\par\end{samepage}\vspace{0.5em}

\begin{samepage}\noindent\hypertarget{ref:sanchez2020}{}\hangindent=1.5em\hangafter=1 Sanchez-Gonzalez A, Godwin J, Pfaff T, 2020. Learning to simulate complex physics with graph networks. Presented at the Proceedings of the 37th International Conference on Machine Learning, PMLR, pp. 8459–8468.\par\end{samepage}\vspace{0.5em}

\begin{samepage}\noindent\hypertarget{ref:simpson2021}{}\hangindent=1.5em\hangafter=1 Simpson, T., Dervilis, N., Chatzi, E., 2021. Machine Learning Approach to Model Order Reduction of Nonlinear Systems via Autoencoder and LSTM Networks. J. Eng. Mech. 147, 04021061. \url{https://doi.org/10.1061/(ASCE)EM.1943-7889.0001971}\par\end{samepage}\vspace{0.5em}

\begin{samepage}\noindent\hypertarget{ref:tancik2020}{}\hangindent=1.5em\hangafter=1 Tancik, M., Srinivasan, P., Mildenhall, B., Fridovich-Keil, S., Raghavan, N., Singhal, U., Ramamoorthi, R., Barron, J., Ng, R., 2020. Fourier Features Let Networks Learn High Frequency Functions in Low Dimensional Domains. Presented at the Advances in Neural Information Processing Systems, pp. 7537–7547. \url{https://doi.org/10.48550/arXiv.2006.10739}\par\end{samepage}\vspace{0.5em}

\begin{samepage}\noindent\hypertarget{ref:wu2024}{}\hangindent=1.5em\hangafter=1 Wu H, Luo H, Wang H, Wang J, Long M, 2024. Transolver: A fast Transformer solver for PDEs on general geometries. Presented at the Proceedings of the 41st International Conference on Machine Learning, PMLR, pp. 53681–53705.\par\end{samepage}\vspace{0.5em}

\begin{samepage}\noindent\hypertarget{ref:yang2025}{}\hangindent=1.5em\hangafter=1 Yang, L.-H., Luo, X.-L., Yang, Z.-B., Nan, C.-F., Chen, X.-F., Sun, Y., 2025. FE reduced-order model-informed neural operator for structural dynamic response prediction. Neural Netw. 188, 107437. \url{https://doi.org/10.1016/j.neunet.2025.107437}\par\end{samepage}\vspace{0.5em}

\begin{samepage}\noindent\hypertarget{ref:zhu2024}{}\hangindent=1.5em\hangafter=1 Zhu, L., Liao, B., Zhang, Q., Wang, Xinlong, Liu, W., Wang, Xinggang, 2024. Vision Mamba: Efficient Visual Representation Learning with Bidirectional State Space Model. \url{https://doi.org/10.48550/ARXIV.2401.09417}\par\end{samepage}\vspace{0.5em}

\end{document}